\documentclass[letterpaper, 10 pt, conference]{ieeeconf} 

\IEEEoverridecommandlockouts 
\usepackage{graphicx}
\usepackage{stackengine}
\usepackage{amsmath}
\usepackage{arydshln}
\usepackage{hhline}
\usepackage{eucal}
\usepackage{cite}
\usepackage{amssymb}
\usepackage{enumerate}
\usepackage[caption=false,font=footnotesize]{subfig}
\usepackage{bm}
\usepackage{color}
\usepackage{multirow}
\usepackage{algorithm2e}

\SetKwComment{Comment}{\scriptsize $\triangleright$\ }{}
\RestyleAlgo{ruled}

\usepackage{graphicx}
\usepackage{svg}
\usepackage{booktabs}
\usepackage{lipsum}
\usepackage{mathtools}

\newtheorem{assumption}{Assumption}

\makeatletter
\renewcommand{\env@matrix}[1][c]{%
  \hskip-\arraycolsep
  \let\@ifnextchar\new@ifnextchar
  \array{*\c@MaxMatrixCols #1}%
}
\makeatother

\DeclareSymbolFontAlphabet{\mathcal}   {symbols}

\usepackage{tikz}
\usetikzlibrary{calc}
\newcommand*\circled[1]{\tikz[baseline=(char.base)]{
    \node[shape=circle, draw, inner sep=1pt, 
        minimum height={\f@size*1.6},] (char) {\vphantom{WAH1g}#1};}}
\makeatother

\title{\LARGE \bf
SPLIT: SE(3)-diffusion via Local Geometry-based Score Prediction \\ for 3D Scene-to-Pose-Set Matching Problems
}

\author{Kanghyun Kim$^{1}$, and Min Jun Kim$^{1}$
\thanks{$^{1}$The authors are with Intelligent Robotic Systems Laboratory, Korea Advanced Institute of Science and Technology, Daejeon, Republic of Korea. {E-mail: {\tt\small \{kh11kim, minjun.kim\}@kaist.ac.kr}}}%
}

\begin{document}

\maketitle
\thispagestyle{empty}
\pagestyle{empty}

\begin{abstract}
To enable versatile robot manipulation, robots must detect task-relevant poses for different purposes from raw scenes. Currently, many perception algorithms are designed for specific purposes, which limits the flexibility of the perception module. We present a general problem formulation called 3D scene-to-pose-set matching, which directly matches the corresponding poses from the scene without relying on task-specific heuristics. To address this, we introduce SPLIT, an SE(3)-diffusion model for generating pose samples from a scene. The model's efficiency comes from predicting scores based on local geometry with respect to the sample pose. Moreover, leveraging the conditioned generation capability of diffusion models, we demonstrate that SPLIT can generate the multi-purpose poses, required to complete both the mug reorientation and hanging manipulation within a single model.
\end{abstract} 

\newcommand{\pose}{\textit{\textbf{H}}}
\newcommand{\posevec}{\textit{\textbf{h}}}
\newcommand{\zerovec}{\textit{\textbf{0}}}
\newcommand{\rot}{\textit{\textbf{R}}}
\newcommand{\trans}{\textit{\textbf{t}}}
\newcommand{\score}{\textit{\textbf{s}}}
\newcommand{\scene}{\textit{\textbf{X}}}
\newcommand{\shape}{\textit{\textbf{Z}}}
\newcommand{\localshape}{{}^{\pose}\bm{z}}
\newcommand{\x}{\mathbf{x}}
\newcommand{\xnoise}{\mathbf{\tilde{x}}}
\newcommand{\sethree}{\text{SE(3)}}
\newcommand{\Exp}{\text{Exp}}
\newcommand{\Log}{\text{Log}}

\section{Introduction} \label{sec:intro}
Detecting task-relevant poses from a raw scene input is a crucial ability for robots to operate autonomously in unstructured environments. For instance, object pose estimation \cite{Wen2023BundleSDFN6, Liu2024DeepLO} focuses on predicting the canonical pose of the target object. In a similar context, object descriptor detection \cite{Simeonov2021NeuralDF, Ryu2023DiffusionEDFsBD} aims to predict object descriptors that annotate object-specific locations in the scene, such as identifying the handle of a mug. Grasp detection algorithms\cite{newbury2023deep, platt2023grasp},
on the other hand, identify graspable poses, regardless of the object's shape or category. These pose detection problems have been actively studied for decades in the fields of robotics and computer vision.

\begin{figure} [tb!]
    \centering
{\includegraphics[width=\linewidth]{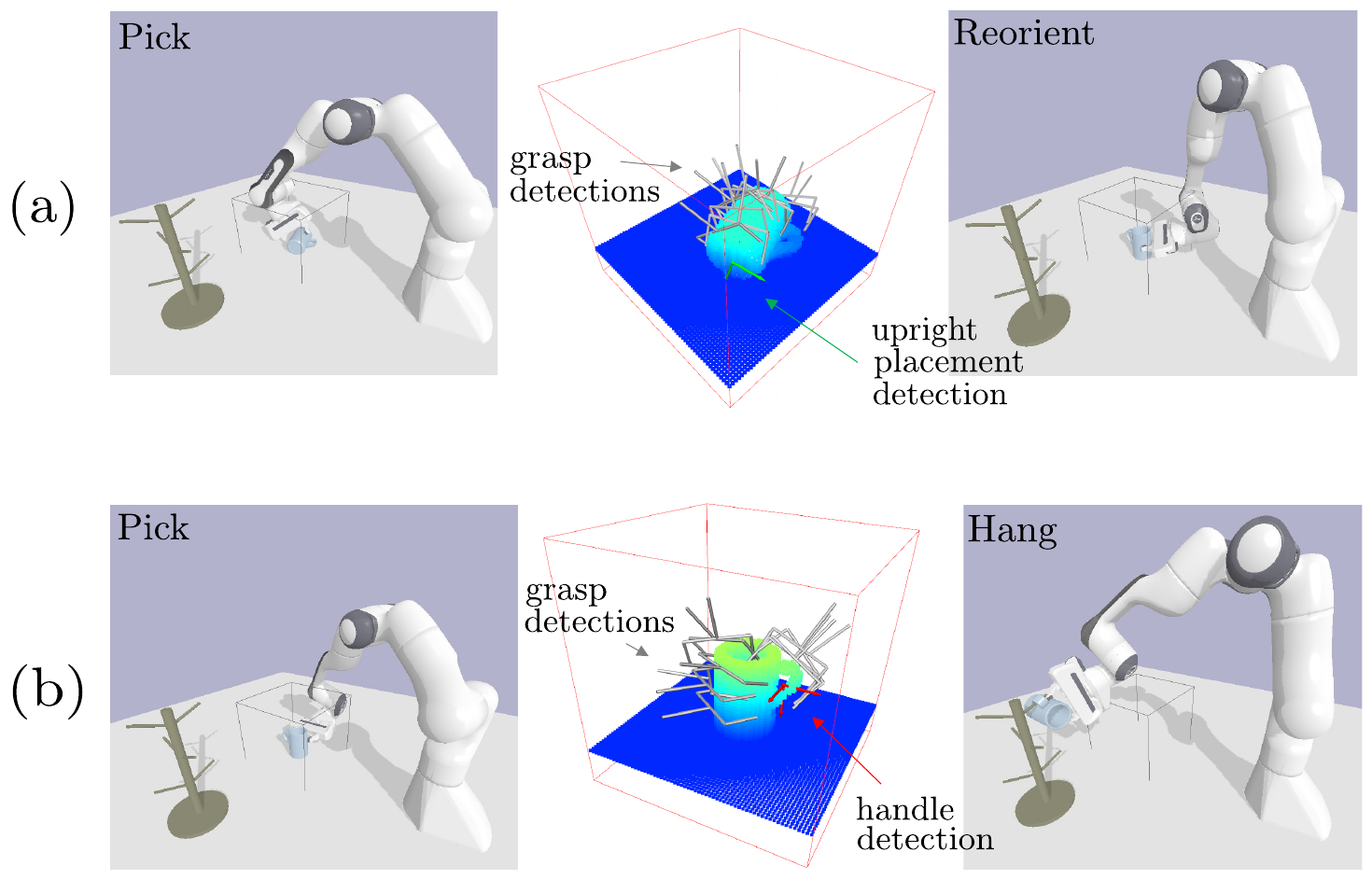}}
    \vspace{-5mm}
\caption{Multiple pose descriptors and grasp detections are required to execute the mug reorientation and hanging task. The robot must determine a stable pose for upright placement, the handle direction for hanging, and grasp candidates for picking. The figure illustrates the subgoal configurations and corresponding poses needed to solve the task: (a) pick-reorient, (b) pick-hang.}
	\label{fig:main}
 \vspace{-5mm}
\end{figure}

Traditionally, each perception algorithm has relied on problem-specific heuristics designed for the particular detection problem at hand. However, as robotic manipulation becomes more complex, multiple perception capabilities are often required simultaneously. For instance, even a simple task like object reorientation (shown in Fig. \ref{fig:main}) requires the robot to determine both how to grasp the object (i.e., grasp detection) and in which pose the object should be placed (i.e., object descriptor detection). As robots are required to perform more complex tasks, relying on single-purpose perception algorithms may limit their adaptability and flexibility. In this paper, we aim to develop a generalizable framework capable of handling a wide range of pose detection problems without the need for problem-specific adjustments. 

To this end, we introduce a new problem called \textit{3D scene-to-pose-set matching} that can encompass various pose detection problems in a single formulation. The formulation assumes that specific geometric features in the scene locally determine the corresponding poses around them, which we call \textit{spatial locality}. For instance, in grasp detection, stable grasp poses are primarily related to the contact geometry. Similarly, detecting an object descriptor for tool use can be determined by the shape of the tool's operational part. This approach offers a more flexible framework, as it does not rely on prior knowledge of specific problems. Instead, it induces the perception module to learn the relationship between geometric features and the corresponding pose sets.

A recent study on SE(3)-diffusion models \cite{urain2022se3dif} provided valuable insights for our 3D scene-to-pose-set matching problem by formulating pose detection as a geometry-based pose generation problem. To generate target poses, the geometry and a pose sample are encoded into a feature vector using an MLP, and the model estimates score functions based on the concatenated featureugh iterative score-based updates, called forward diffusion, the pose samples converge to the desired poses. However, we observed that this approach struggles when complex geometry such as scene input is considered. This limitation arises from the model’s lack of spatial reasoning. When geometry and the pose query are compressed into a single latent vector, the model fails to capture local geometric details and cannot leverage inductive biases like SE(3)-equivariance. This issue was also discussed in 3D reconstruction tasks \cite{Peng2020convoccnet, chibane20ifnet}.

The goal of this paper is to develop an SE(3)-diffusion model that generates SE(3) poses relevant to a given scene, aiming to address various pose detection tasks using a single framework. We call our model \textbf{SPLIT}: \textbf{S}core \textbf{P}rediction with \textbf{L}ocal Geometry \textbf{I}nformation for Rigid \textbf{T}ransformation Diffusion. Instead of encoding global geometry and a sample pose, SPLIT encodes the local geometric context with respect to the sample pose. This enhances model efficiency, as 1) predicting poses with a localized view of the geometry is simpler than using the full geometry and 2) the same local geometry encoder can be applied to all SE(3) pose samples in the scene, naturally exploiting symmetry.
To summarize, the contributions of this paper are as follows. First, we present a new problem formulation called 3D scene-to-pose-set matching to tackle a broad class of pose detection tasks within a unified framework (Sec. \ref{subsec:probdef}). We identify the concept of spatial locality as a key insight for solving this problem effectively. Second, as a solution, we propose SPLIT, a novel SE(3)-diffusion model (Sec. \ref{subsec:split}), which utilizes spatial locality in its architecture. Using SPLIT, we demonstrate that the mug reorientation and hanging tasks (Fig. \ref{fig:main}), which require solving multiple pose detection problems, can be handled with a single model.

\section{Related Work}
\subsection{Learning-based Grasp Detection Approaches}
Grasp detection is a fundamental area of research in robotic manipulation \cite{bicchi2000robotic, bohg2013data}. Early methods employed mathematical models for friction and geometry to plan grasps \cite{murray2017mathematical, Miller2004GraspitAV}. More recent research has focused on learning-based approaches to enhance generalization for unseen objects and enable fast grasp queries \cite{newbury2023deep, platt2023grasp}.

These methods typically take raw inputs, such as RGB or depth images, and output a finite set of grasp candidates. However, as directly regressing SE(3) poses presents significant challenges, many previous studies have utilized heuristics to address this issue. One successful approach is to utilize discretized candidates, such as voxels in a 3D grid or points in point clouds \cite{breyer2021volumetric, sundermeyer2021contact, gou2021rgb, ni2020pointnet++}. The model first evaluates the quality of each candidate, then regresses the remaining degrees of freedom for the selected points. On the other hand, there is another approach to solving this as a generation problem. From this perspective, generating accurate poses that match complex geometries, such as a scene, is challenging. Therefore, object segmentation is typically performed first, and the algorithm generates grasps based on the object-centric shape embedding \cite{Mousavian20196dofgrasp, urain2022se3dif}.

\subsection{Learning-based Object Description Approaches}
Due to the shape variance within an object category, object pose estimation is often insufficient for manipulation. For example, even if the accurate canonical pose of a mug is known, the position of the handle may vary depending on its shape. In this context, object description methods focus on extracting task-relevant spatial features from the scene at the category level. A common approach is to detect keypoints from an image and use them to derive the 6-DoF poses\cite{Manuelli2019kPAMKA, Gao2021kPAM2F, Florence2018DenseON}. However, these methods depend heavily on the keypoint prediction accuracy, which can significantly reduce task success rates. For this, recent methods directly infer the 6-DoF pose without relying on keypoints \cite{Simeonov2021NeuralDF, Ryu2023DiffusionEDFsBD, Ryu2022EquivariantDF}.

\subsection{SE(3)-diffusion Models}
Diffusion models \cite{ho2020ddpm, Song2019ncsn} have recently shown strong performance in generating realistic data across various fields. Several researchers have adapted this ability to pose generation problems in robotic manipulation. Urain et al. \cite{urain2022se3dif} introduced an SE(3)-diffusion model that generates multiple grasp poses based on object shape. However, the model simply compresses the object geometry and the sample pose into a single vector, which limits its capability to handle complex spatial relationships. To address this, Singh et al. \cite{Singh2024Constrained6G} employed a convolutional multi-plane encoder \cite{Peng2020convoccnet} to handle part-constrained pose generation from an object point cloud. The most relevant work to ours is Diffusion-EDF \cite{Ryu2023DiffusionEDFsBD}, which takes full-view scene input and outputs an SE(3) pose descriptor for manipulation. For this, the method uses an SE(3)-equivariant graph neural network encoder, which exploits symmetry in geometric reasoning. In contrast, our method utilizes a lighter CNN-based architecture and demonstrates applicability even with partial-view inputs.

\section{Preliminaries}

\subsection{Score-based Generative Models}
Diffusion models generate new data by progressively denoising samples from pure noise \cite{ho2020ddpm}. This process can also be interpreted through score-based generative modeling \cite{song2020scoresde}, which involves estimating the score, or the gradient of the log data density function. Since score-based sampling techniques\cite{parisi1981correlation, neal2012mcmc} can be used to sample from the target distribution, the primary objective of this framework is to accurately learn the score function using a neural network.

Since the true score function is unknown, Vincent et al. \cite{Vincent2011ACB} proposed denoising score matching (DSM) for training a network. In this method, a noise kernel $q_\sigma(\tilde{x}|x) = \mathcal{N}(\tilde{x}; x, \sigma^2 \mathbf{I})$ is applied to produce a perturbed data distribution $p_\sigma(\tilde{x})$. Then, instead of learning the exact score function directly, the model $\bm{s_\theta}$ is trained to approximate the score of the perturbed distribution by minimizing the DSM loss:
$$\mathcal{L}_{DSM}=\frac{1}{L} \sum_{k=0}^L \mathbb{E}_{x, \tilde{x}} [||\bm{s_\theta}(\tilde{x}) - \nabla_{\tilde{x}} \log q_\sigma(\tilde{x}|x)||^2_2].$$
It is important to note that the choice of noise scale presents a tradeoff: smaller perturbations yield more accurate score estimates, but the model's coverage is reduced because the training relies on perturbed data.

To mitigate this issue, a noise-conditioned score model $\bm{s_\theta}(\tilde{x}, \sigma_k)$ is introduced\cite{Song2019ncsn}. During training, known as forward diffusion, the data is perturbed with an increasing noise schedule from $k{=}0$ to $k{=}T$. At the initial noise level $\sigma_0$, the distribution $p_{\sigma_0}(\tilde{x})$ is close to $p_{data}$. By the final noise level $\sigma_T$, the distribution $p_{\sigma_T}(\tilde{x})$ contains no useful information about the original data. The same DSM loss can be applied to train the noise-conditioned score network $\bm{s_\theta}(\tilde{x}, \sigma_k)$.

For sampling, known as inverse diffusion, the predicted score at time $k$ is used for a score-based sampling technique, such as Langevin dynamics. Starting with an initial sample at time $T$, $\tilde{x}_T \sim p_{\sigma_T}(x)$, the model iteratively updates the sample using the following dynamics:
$$\x_{k-1} = \x_{k} + \frac{\alpha_k}{2}\mathbf{s}_{\theta}(\x_{k}, \sigma_k) + \sqrt{\alpha_k} \mathbf{\epsilon}, \quad \mathbf{\epsilon} \sim \mathcal{N}(\mathbf{0}, \mathbf{I}),$$
where $\alpha_k$ is a time-dependent step size. By updating the sample until $k{=}0$, we obtain a sample from the original data distribution.

\subsection{SE(3) Lie group}\label{subsec:liegroup}

In robotics, representing the pose of objects, robot links, and joints is essential. A commonly used representation is the $4\times4$ homogeneous transformation matrix, $\pose = \begin{bsmallmatrix} \rot & \trans  \\  \mathbf{0} & 1 \end{bsmallmatrix}$, where $\trans$ is a position vector, and $\rot$ is a rotation matrix. These matrices form a differentiable manifold known as $\sethree$ \cite{Sol2018microlie}.

For a pose $\pose$ moving on the $\sethree$ manifold, its velocity is represented in the tangent space $T_\pose \sethree$, which is isomorphic to a Euclidean space, $\mathbb{R}^6$. The tangent space at the identity pose $\mathcal{E}$ is called the Lie algebra $\mathfrak{se}(3):= T_{\mathcal{E}}\sethree$, and there exists a mapping between the Lie group $\sethree$ and the Lie algebra $\mathfrak{se}(3)$, known as the exponential and logarithmic maps: $\exp:\mathfrak{se}(3)\rightarrow \sethree$ and $\log:\sethree\rightarrow \mathfrak{se}(3)$. Consequently, we extend this mapping to the isomorphic $\mathbb{R}^6$ space, referred to as $\Exp: \mathbb{R}^6\rightarrow\sethree$ and $\Log: \sethree\rightarrow\mathbb{R}^6$. These maps allow any element of $\sethree$ to be parameterized as a vector. In this paper, we call it the \textit{global parametrization}, as $\sethree$ components are located in the global tangent space.

Similarly, \textit{local parametrization} with respect to a given pose $\pose$ can be defined. For a pose $\pose'$, the mapping between $\sethree$ and the local tangent space at $\pose$ is given by
$${}^{\pose}\Exp(\posevec'):=\pose\Exp(\posevec') \text{  and  } {}^{\pose}\Log(\pose')=\Log(\pose^{-1}\pose'),$$
where $\posevec'$ is a vectorized component of $\pose'$ in the tangent space $T_\pose \sethree$.

Uncertainty in $\sethree$ can be represented by defining a Gaussian distribution in the tangent space. Given a mean pose $\bar{\pose}$, a set of noisy poses can be generated by sampling zero-mean perturbation vectors from the tangent space $T_{\bar{\pose}}SE(3)$ and multiplying them to $\bar{\pose}$. The probability distribution can be expressed as:
$$q(\pose|\bar{\pose}, \Sigma)\propto \exp \left(-0.5 \, \| {}^{\bar{\pose}}\Log(\pose) \|^2_{\Sigma^{-1}} \right),$$ 
where $\Sigma=\sigma\bm{I}$ is a covariance matrix. We denote this distribution as $q_{\sigma}(\pose|\bar{\pose})$.

Finally, we define the gradient of a function in $\sethree$. As an $\sethree$ pose can be mapped to a vector, the gradient represents the rate of change of a function in the parameterized vector space. When local parametrization at $\pose$ is used, the gradient of $f(\pose)$ at $\pose'$ is given by:
$$\left.\frac{{}^{\pose} D f(\pose)}{{D\pose}}\right\rvert_{\pose=\pose'} = \left.\frac{\partial f(\posevec)}{\partial \posevec}\right\rvert_{\posevec=\posevec'} \quad \posevec' = {}^{\pose}\Log(\pose'),$$
where $\posevec$ is the parameterized pose vector in the tangent space $T_{\pose} \sethree$. Throughout this paper, we denote this operation as ${}^{\pose}\nabla f(\pose')$. In practice, this gradient is computed using automatic differentiation. For more details, refer to \cite{Sol2018microlie}.

\section{Method}

\subsection{3D scene-to-pose-set matching: Towards a Generalizable Pose Detection Formulation}\label{subsec:probdef}
We aim to define a more flexible problem, which we refer to as 3D scene-to-pose-set matching. This problem adheres to the following key assumptions:
\begin{assumption}
The target pose has a local relationship with specific parts of the scene. We refer to this relationship as the \textit{local geometric context}. This context captures the implicit connection between the scene and the target pose.
\end{assumption}
\begin{assumption}
The same local geometric context consistently leads to the same target poses. This means that if the local geometric context is identical, two different locations can generate similar target poses relative to their positions.
\end{assumption}

To realize this abstract concept, we assume that a neural network $\mathcal{F}_{\phi}$ can encode the local geometric context: 
$$\mathcal{F}_{\phi}(\pose, \scene) = \localshape,$$
where the inputs of the network are a pose query $\pose \in \sethree$ and a scene $\scene$. The output, $\localshape \in \mathbb{R}^n$, is a latent vector that encodes local geometric context with respect to the frame of the pose query. Intuitively, the network learns to extract the task-relevant information around the given pose $\pose$ within the entire scene.

\begin{figure} [tb!]
    \centering
	{\includegraphics[width=\linewidth]{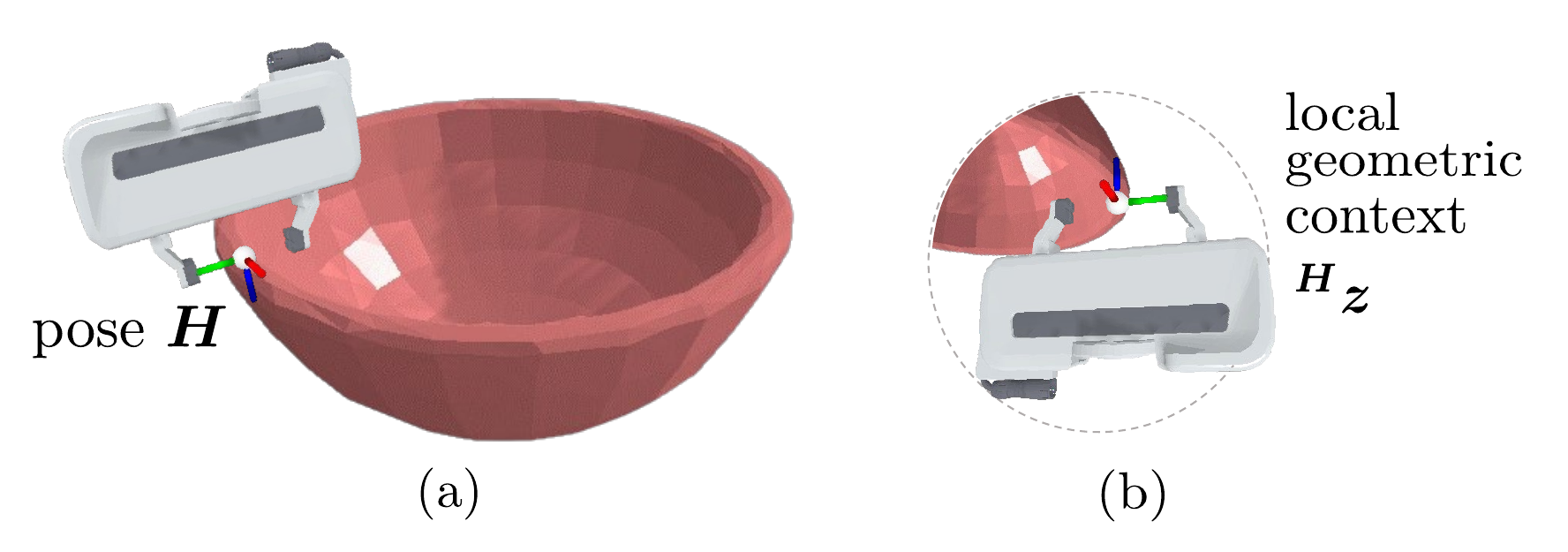}}
    \vspace{-7mm}
    \caption{A visual explanation of the spatial locality assumption of our problem. For example, in a grasp detection task, the success probability at a given pose is induced by the local geometric context surrounding it.}
	\label{fig:assumption}
 \vspace{-5mm}
\end{figure}

Building on this, we mathematically represent the concept of \textit{spatial locality} in our problem. The success probability of the pose detection task depends solely on the local geometric context around the pose and is independent of the global scene. This relationship can be expressed as: 
\begin{equation}\label{eq:assumption} p(\text{succ} | \pose, \scene) = p(\text{succ} | \localshape), \end{equation} 
where $\text{succ}$ denotes the success event of the pose detection task. Fig. \ref{fig:assumption} illustrates this idea in the context of grasp detection. It shows that the success probability is unaffected by the current pose or unrelated geometries, as long as the task-related local geometric context $\localshape$ remains constant.

\subsection{SPLIT: Sample-frame Score Prediction based on the Local Geometry}\label{subsec:split}

We aim to address the 3D scene-to-pose-set matching problem with an efficient SE(3)-diffusion model that leverages spatial locality. The key intuitions from Sec. \ref{subsec:probdef} are as follows:
\begin{itemize} 
\item \textbf{Locality}: By spatial locality of the problem, global geometry becomes unnecessary for score prediction.
\item \textbf{Symmetry}: Instead of feeding both geometry and a pose query into the score prediction network, we can use a geometry feature canonicalized with respect to the pose query. \end{itemize} 

From a generative modeling perspective, pose detection involves sampling from the distribution of successful poses, $\pose \sim p(\pose | \text{succ}, \scene)$. Let $\score(\pose, \scene)$ be the score function of our target distribution. Using \eqref{eq:assumption}, we decompose the score function: 
\begin{align} 
\score(\pose, \scene) &= \nabla \log p(\pose|\text{succ}, \scene) \\ 
& = \nabla \log \frac{ p(\text{succ}| \pose, \scene)p(\pose|\scene)}{p(\text{succ}|\scene)} \\ 
& = \nabla \log p(\text{succ}| \pose, \scene) \\ 
& = \nabla \log p(\text{succ}| \localshape). 
\end{align} 
First, note that we omit the notation of parametrization needed to define a gradient (Sec. \ref{subsec:liegroup}), as any parametrization is valid in this context. In (3), two terms are excluded: (i) $p(\pose|\scene)$ is treated as a uniform distribution since no task-specific prior knowledge is used for generating pose candidate $\pose$ from the scene $\scene$, and it is therefore removed by the gradient. (ii) $p(\text{succ}|\scene)$ is independent of $\pose$, so it is also removed. From this, we conclude that, rather than redundantly using global scene knowledge, it is sufficient to derive score prediction from the local geometric context $\localshape$.

The choice of parameterization for the gradient is also crucial. Here, we denote $\log p(\pose|\text{succ}, \scene):=f(\pose)$ for convenience. If global parameterization is used, the score function is represented as
$${}^{\mathcal{E}}\nabla f(\pose)=\frac{\partial f(\posevec)}{\partial \posevec}=\frac{\partial f(\Log(\pose)}{\partial \posevec},$$
which indicates that the score prediction depends on the current pose $\pose$. In contrast, if local parameterization at $\pose$ is used:
$${}^{\pose}\nabla f(\pose)=\frac{\partial \score(\posevec)}{\partial \posevec}=\frac{\partial f({}^{\pose}\Log(\pose))}{\partial \posevec},$$
and since ${}^{\pose}\Log(\pose)=\bm{0}$, this implies that the current pose is not required for score prediction if the predicted score is represented on the tangent space $T_\pose\sethree$.

In conclusion, our score model $\score_\theta$ predicts the SE(3) score, parametrized with respect to the sample frame. For the input, a local shape encoder $\mathcal{F}_\phi$ encodes the latent local geometric context $\localshape$ from a given scene $\scene$:
$$\mathcal{F}_\phi(\pose, \scene)=\localshape, \quad \score_\theta(\localshape) \simeq {}^{\mathcal{\pose}}\nabla \log p(\pose | \text{succ}, \scene)$$ 

This structure provides two key advantages: 1) The score prediction $\score_\theta(\localshape)$ is performed over simpler local geometries $\localshape$ rather than the global geometry. 2) Since the network input $\localshape$ is a geometric feature canonicalized in the sample pose frame, the architecture leverages symmetry in the score prediction\footnote{Although we achieve SE(3)-equivariance in the score prediction, the overall architecture partially leverages this property, as the convolutional neural network used as our geometry encoder is limited to translation equivariance.}.

\begin{figure}[t]
\centering
  \includegraphics[width=\linewidth]{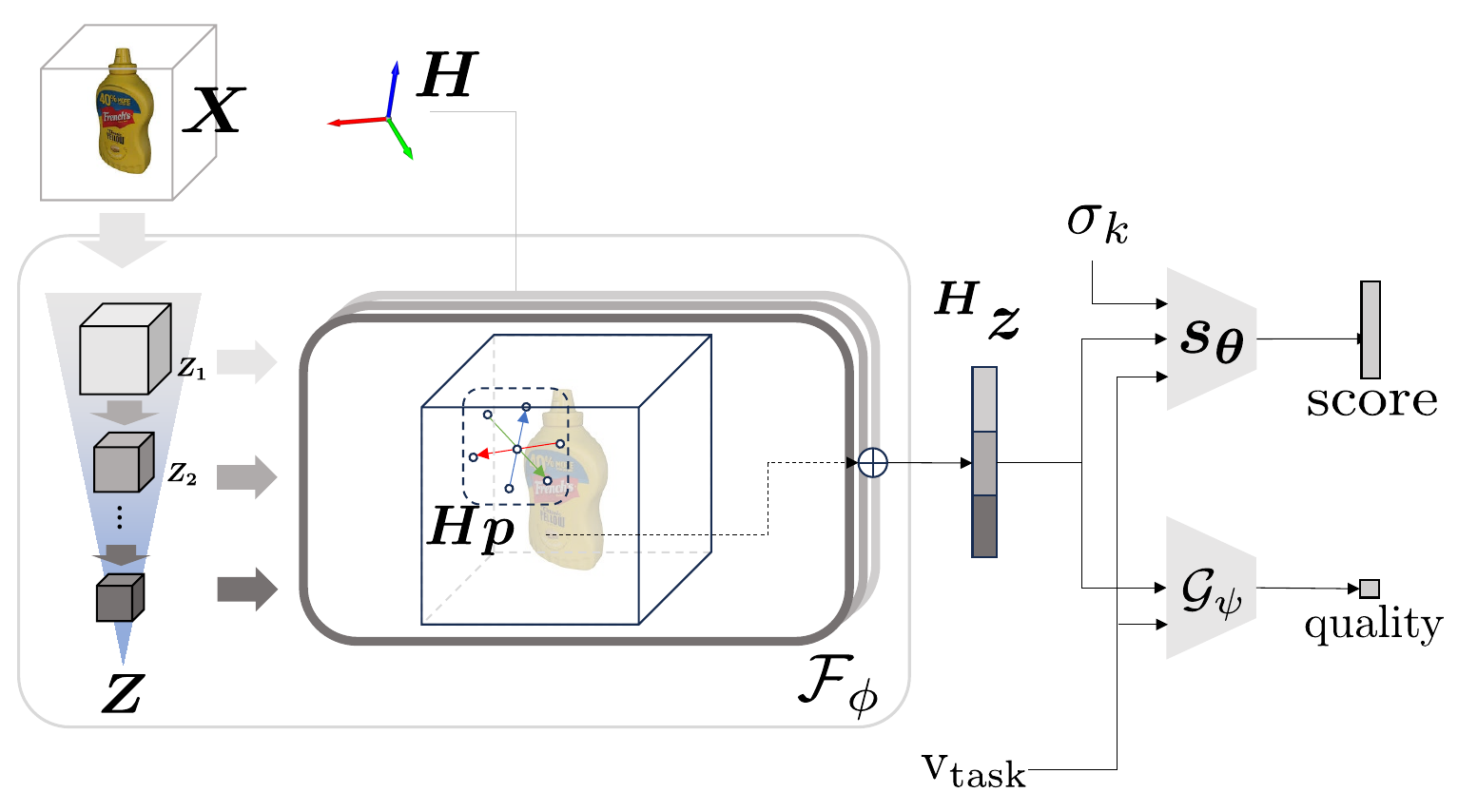}
  \vspace{-7mm}
  \caption{The network architecture of SPLIT. To extract the local geometric context $\localshape$ from the multi-scale feature grids $\shape$, a point kernel is transformed by the sample pose to interpolate the local features at the points.}
  \label{fig:arch}
  \vspace{-4mm}
\end{figure}

\subsection{Network Architecture for SPLIT} \label{subsec:impl}
The overall architecture of our network is depicted in Fig. \ref{fig:arch}. A scene input $\scene$ is a 3D occupancy grid. To generate this grid, single or multi-view depth images are first converted into a point cloud, and the occupancy of points within a predefined workspace grid is counted.

To encode the local geometric context $\localshape$ from scene $\scene$, we adopt an architecture similar to IF-Nets \cite{chibane20ifnet}. First, multiple downscaling CNN blocks are applied to generate multi-scale feature grids $\shape = (\bm{Z_1}, \bm{Z_2}, \dots)$. To encode $\localshape$ from the feature grids $\shape$, the pose $\pose$ is used to transform a predefined point kernel $\bm{p}$ to obtain interpolated feature values at these points, resulting in $\localshape = (\bm{Z_1}(\pose \bm{p}), \bm{Z_2}(\pose \bm{p}), \dots)$. The point kernel $\bm{p}$ consists of seven points: one central point and six points located at a distance $d$ along the principal axes.

Finally, the concatenated local feature vector $\localshape$ is fed into two 8-layer MLPs: score prediction network $\bm{s_\theta}$ and sample evaluation network $\mathcal{G}{\psi}$. As in \cite{Mousavian20196dofgrasp}, the sample evaluation network is trained to distinguish original samples from negative samples (e.g., grasp failure poses) and/or noisy samples generated during the training process. This network is used for rejecting low-quality samples and prioritizing generated samples. 

In addition, we leverage the conditional generation capability of diffusion models to handle multiple pose generation tasks within a single model. For this, both networks utilize a FiLM architecture for conditioning inputs \cite{Perez2017FiLMVR}. A one-hot vector $\text{v}_{\text{task}}$ representing the target task is provided as an additional condition. Furthermore, for noise-conditioning of the score prediction network, the noise scale at time $k$, $\sigma_k$, is incorporated using a Gaussian Fourier encoder \cite{Tancik2020FourierFL}, which allows the model to encode the noise scale as a frequency-based feature.

\subsection{Sample-frame SE(3)-diffusion}\label{subsec:sample-frame-diff}
Here, we define a sample-frame $\sethree$ diffusion model that utilizes a locally parameterized score function. Note that it differs from the SE(3)-diffusion in previous works, as they used a globally parameterized score function \cite{urain2022se3dif, Singh2024Constrained6G, Ryu2023DiffusionEDFsBD}.

During training, a $\sethree$ noise kernel with various noise scales $\sigma_k$ is applied to perturb the original data samples. This is done by sampling perturbation vectors $\text{v} \sim \mathcal{N}(\bm{0}, \sigma_k^2 \bm{I})$ and perturbing the original data as $\tilde{\pose} = \pose \Exp(\text{v})$. After perturbing the original data, latent local geometric features are then computed for each noisy pose, ${}^{\tilde{\pose}}\bm{z} = \mathcal{F}_\phi(\tilde{\pose}, X)$.

Our DSM loss objective is to match the score predictions with the locally parameterized target scores:
$$\mathcal{L}_{DSM}=\frac{1}{L} \sum_{k=0}^L \mathbb{E}_{\pose, \tilde{\pose}} [||\bm{s_\theta}(\bm{{}^{\tilde{\pose}}z}, \sigma_k) - {}^{\tilde{\pose}}\nabla \log q_{\sigma_{k}}(\tilde{\pose}|\pose)||^2_2],$$
where $q_{\sigma_k}$ is a Gaussian distribution defined in SE(3) (Sec. \ref{subsec:liegroup}).
In practice, the target scores ${}^{\tilde{\pose}}\nabla \log q_{\sigma_{k}}(\tilde{\pose}|\pose)$ are computed using automatic differentiation. 

Pose generation is achieved by iteratively updating initial pose samples $\pose_{T}$ through simulating Langevin dynamics from $k=T$ to $k=0$. Practically, the initial pose sample is drawn from uniformly random poses in the scene\footnote{This is because: 1) applying a noise kernel with a large noise scale converges to a uniform distribution in rotation due to the compact nature of SO(3), and 2) the score outside the scene cannot be defined.}. At each step, sample-frame Langevin dynamics is applied as follows:
$$\pose_{k-1} = \pose_{k} \Exp(\frac{\alpha_k}{2}\bm{s_{\theta}}(\localshape, \sigma_k) + \sqrt{\alpha_k} \bm{\epsilon}), \quad \bm{\epsilon} \sim \mathcal{N}(\mathbf{0}, \mathbf{I}).$$
Note that $\frac{\alpha_k}{2}\bm{s_{\theta}}(\localshape) + \sqrt{\alpha_k} \bm{\epsilon}$ represents a one-step Langevin update in the tangent space $T_{\pose_k}\sethree$.

\section{Implementation Details}

\begin{figure}[t]
\centering
  \includegraphics[width=0.95\linewidth]{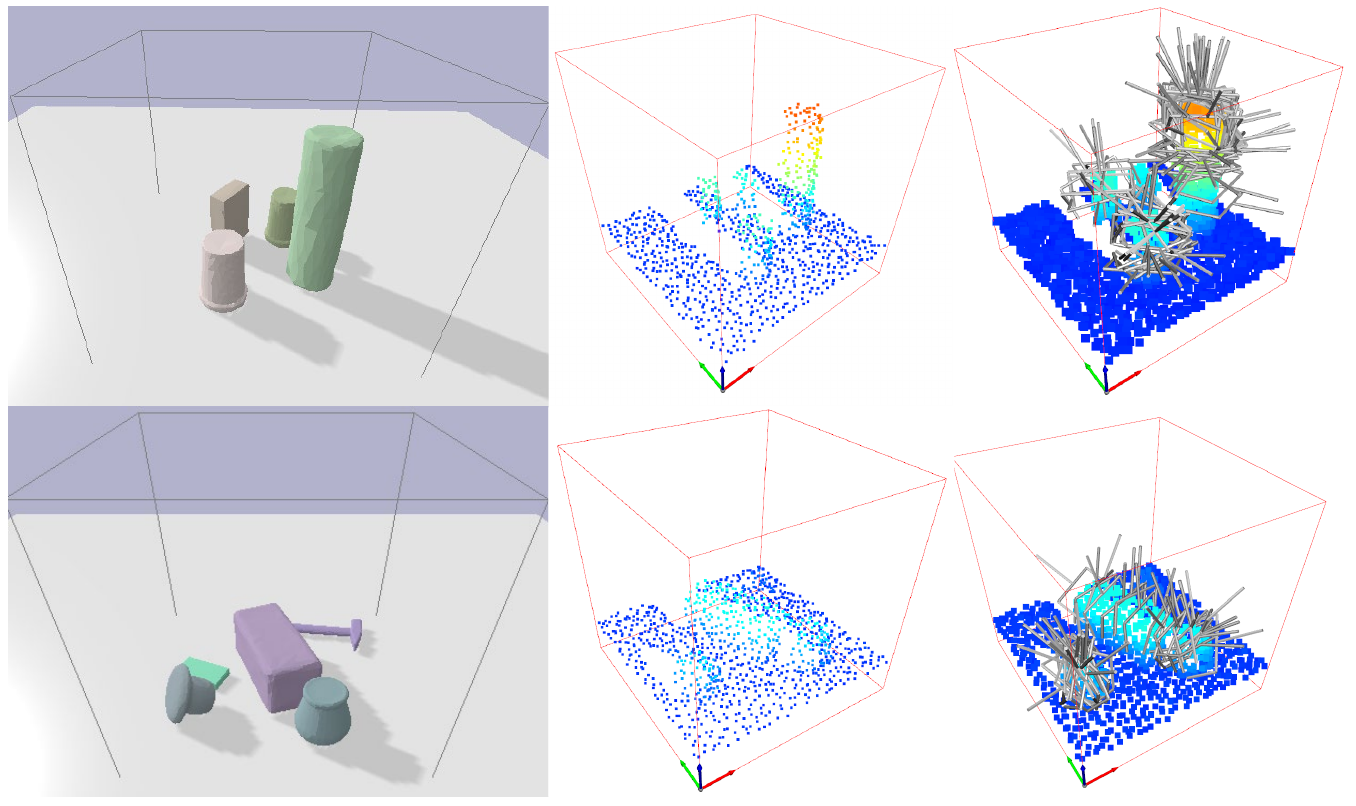}
  \vspace{-2mm}
  \caption{Packed and pile scenes from \cite{breyer2021volumetric}, along with grasp generation results. A point cloud is obtained from a single-view depth image and converted into an occupancy grid for input.}
  \label{fig:vgnenv}
  \vspace{-4mm}
\end{figure}

\noindent\textbf{Synthetic Data Generation Protocol:} The training datasets for all experiments were generated through simulation. The generation protocol is as follows: first, target pose candidates were generated at the object level using antipodal sampling for grasp detection \cite{Eppner2019ABW}, or through manual annotation for the object description task. These candidates were then attached to each object and placed in the scene. 

The feasibility of grasp poses was further validated through physics simulations. The feasibility check included collision detection between the gripper and the ground, as well as ensuring that the angle between the approach axis and the z-axis was less than 90 degrees. Additionally, grasp visibility was checked by calculating the point where the grasp approach vector intersects the object surface and ensuring the distance between this point and the point cloud was less than 0.005. Grasp candidates that failed these checks were used as negative samples for training the evaluator. 

The scene input was captured using a single-view depth camera. To minimize the sim-to-real gap, we applied farthest point downsampling to 2000 points on the point cloud obtained from the depth image, and Gaussian noise with a standard deviation of 0.005 was added to the simulated point cloud. The point cloud was converted into a $64{\times}64{\times}64$ occupancy grid, which has an edge length of 0.3m. For each experiment, a set of 200 train objects with different scales was used to generate 30,000 training scenes.

\noindent\textbf{Training Details:} The geometry encoder was based on the IF-Net architecture, using five CNN layers with downscaling by a factor of 2 at each step. For the multiscale feature grids, the input, output, and the 1st, 2nd, and 3rd intermediate grids were utilized. The displacement for the point kernel was set to 8cm, corresponding to the gripper width in the actual workspace. The overall network loss was a combination of DSM loss for the score prediction network and focal loss for the evaluation network, $\mathcal{L}_{DSM} + 0.1 \mathcal{L}_{focal}$, with focal loss parameters set to $\alpha{=}0.25$ and $\gamma{=}2$. To train the evaluation network, we determine a time threshold $t_{th}{=}0.3$, and noisy samples generated from a noise schedule after $t_{th}$ are considered negative samples. The training was conducted using the Adam optimizer with a learning rate of 0.001. For the diffusion process, the noise schedule ranged from 0.02 to 0.1. A total of 1000 timesteps were used for training, while 100 timesteps were employed for sampling.

\section{Experiments}

\subsection{Scene-based Grasp Generation Evaluation}
\begin{table}
\centering
\caption{Evaluation Results on Scene-based Grasp Generation}

\begin{tabular}{ccccc} 
\hline\hline
Method                                           & \multicolumn{2}{c}{Packed} & \multicolumn{2}{c}{Pile}  \\
                                                 & GSR (\%)  & DR (\%)        & GSR (\%)  & DR (\%)       \\ 
\hline 
VGN~\cite{breyer2021volumetric} & 76.37±3.3& 79.26±3.7& 47.60±3.7& 31.06±3.57\\
GIGA~\cite{jiang2021giga}       & 84.04±3.5& 83.48±2.8& 56.83±5.7& 39.42±5.52\\
SE(3)-Dif~\cite{urain2022se3dif}  & 35.34±2.8& 28.35±3.6& 10.62±2.2& 4.94±1.1\\ 
\hline
SPLIT (Ours)                 &           \textbf{89.01±2.3}&                \textbf{86.63±4.6}&           \textbf{69.00±6.3}&               \textbf{47.59±4.9 }\\
\end{tabular}
\label{tab:grasp}
\vspace{-6mm}
\end{table}

In the first experiment, we aimed to validate the grasp pose generation capability in complex scenes, using the clutter removal task introduced in \cite{breyer2021volumetric}. The task included two scene types: packed and pile. Each scene is generated in simulation using five unseen objects in random poses. Given a single-view depth image, a grasp pose was inferred by the model and executed in a physics simulation. Grasp attempts continued each round until one of three termination criteria was met: 1) all objects were removed, 2) no feasible grasps remained, or 3) two consecutive grasp attempts failed. Performance was measured using Grasp Success Rate (GSR) and Declutter Rate (DR). GSR is the ratio of successful grasps to total trials, and DR is the ratio of successful grasps to the total number of objects in the scene. Test scenes were generated using five objects from unseen test sets, and each model was tested over 40 simulation rounds across five random seeds, for a total of 200 scenes.

We employed three baselines: VGN \cite{breyer2021volumetric}, GIGA \cite{jiang2021giga} (with pre-trained parameters from the GIGA authors), and SE3-Dif \cite{urain2022se3dif}. SE3-Dif and our model were trained on the same partial point cloud dataset, but unlike the original implementation, we omitted the signed distance prediction loss, which is challenging to compute in 3D scenes.

Table \ref{tab:grasp} shows the simulation comparison results. Our method performed comparably to existing grasp pose detection algorithms. Notably, the baseline SE(3)-Dif model did not perform well across both scene types. Effective matching of grasp poses to the scene requires attention to fine geometric details, which is difficult when using global geometry encoding. In contrast, our approach extracts local geometric context with a point kernel, preserving finer details and enabling more efficient score prediction for each sample. The same experiment was also conducted in real-world scenarios. For more details, please refer to the attached video.

\subsection{Multi-purpose Pose Generation for Mug Reorientation and Hanging Manipulation}

\begin{figure}[t]
\centering
  \includegraphics[width=0.95\linewidth]{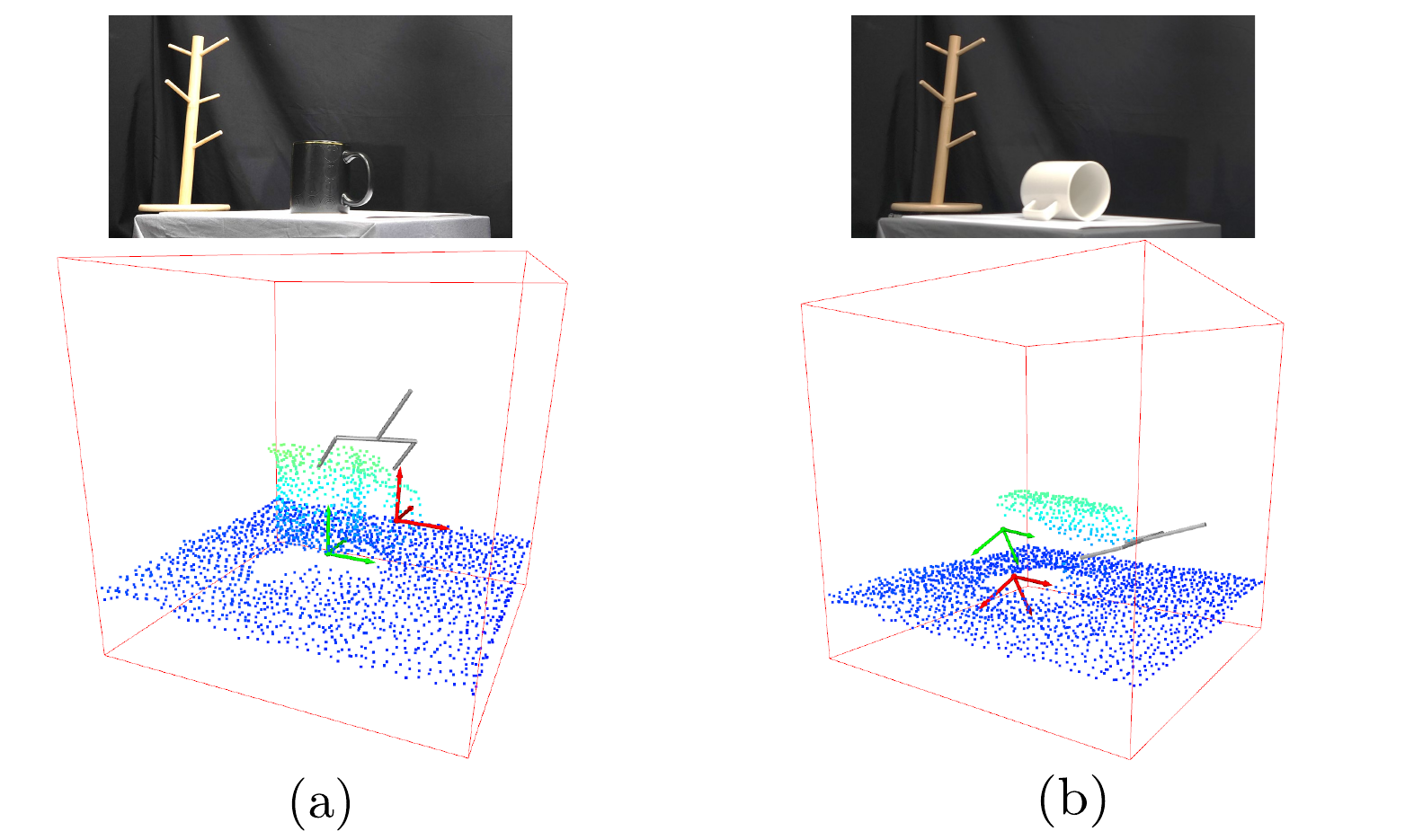}
  \vspace{-2mm}
  \caption{Two real-world examples of multi-purpose pose detection with the highest probability are presented. (a) The mug is positioned in the canonical pose. (b) Even though the mug’s handle is not visible in the point cloud, SPLIT infers implicit information (i.e., the handle is located on the hidden side in the stable pose) from the scene-pose dataset.}
  \label{fig:mug}
  \vspace{-5mm}
\end{figure}

We conducted the second experiment to evaluate the generalized scene-to-pose-set mapping capability of our model by performing a mug hanging task, as shown in Fig. 1. A mug was initially placed in a random orientation around the origin. From the captured scene, the model aimed to detect three types of poses: graspable poses, a handle pose, and an upright placement pose. 

In certain initial mug poses, a kinematically feasible solution for hanging may not exist. To deal with such cases, if the planner fails to find a solution, the robot attempts an intermediate subgoal of reorienting the mug to an upright position. Cases where the mug was upside down were excluded, as no grasp solution exists for such orientations. All pick-and-place operations are executed using pre-grasp and pre-placement poses. To avoid ambiguity in the mug's pose within the scene, two depth images were captured from different angles: one directly perpendicular to the workspace and another rotated 45 degrees. Mug shapes were manually selected from the ShapeNet dataset \cite{Chang2015ShapeNetAI}, with 25 mugs used for training and 5 reserved for testing. We compared our approach to executions based on ground truth perception, which is a pre-calculated pose set for the object. The evaluation was based on the success rate of the overall task. Additionally, to isolate the impact of the reorientation task, we specifically measured its success rate separately. 
The experiment was conducted 40 times for both the mug reorientation and hanging tasks. In the ground truth results, the mug hanging task achieved a success rate of 77.5\%. It is important to note that even with knowledge of the ground truth poses, execution often fails due to small pose errors during grasping or reorienting motion, or the lack of a feasible plan. When using the predicted poses by SPLIT, the success rate was 70\%, yielding a relative success rate of 90.3\%. Additionally, the success rate of the mug reorientation with SPLIT was 92.5\%.

We identified two main factors contributing to task failures in perception. First, the model did not favor certain robust grasp poses. For example, selecting a shallow grasp on the rim often caused orientation errors during the motion due to a lack of robustness. This is expected, as our method predicts target poses solely based on scene geometry and does not account for task-specific heuristics, such as robustness prediction based on the center of mass \cite{mahler2017dex}. However, we believe that task-specific pose preferences could be improved by incorporating them into scene-pose dataset generation. Second, when the mug handle was completely invisible in the scene, incorrect pose predictions led to manipulation failures. This challenge arose from the lack of explicit handle information, as shown in Fig. \ref{fig:mug}(b). Nevertheless, our method learned priors from the dataset, enabling it to infer handle positions without heuristics, and in some cases, it predicted feasible poses. This highlights one of the key strengths of our approach: its ability to learn the implicit relationship between scene geometry and poses from the dataset without relying on hand-crafted rules. Therefore, we expect that, as the model is trained with larger and more diverse datasets, this end-to-end approach will become even more scalable, similar to recent successes of large models.

\section{Conclusion}
This paper introduced 3D scene-to-pose-set matching, a novel formulation to address a broad range of pose detection tasks within a unified framework. By identifying spatial locality as a key factor for improving efficiency, we developed SPLIT, a score-based SE(3)-diffusion model that leverages this concept. The mug reorientation and hanging experiment demonstrated SPLIT's ability to generate multi-purpose poses with a single model, showing performance comparable to ground truth in terms of task success rate.

We also identify several limitations. SPLIT struggles with out-of-distribution scenes, a common issue in end-to-end models. It is expected that similar to advances in image-based models, larger datasets or domain randomization could mitigate this issue. We also observed that addressing real-world uncertainties during execution is crucial for improving success rates. Exploring how feedback can be incorporated to enhance success rates, as in \cite{Gao2021kPAM2F}, is an important direction for our future work.

\bibliographystyle{myIEEEtran.bst}
\bibliography{IEEEabrv,my.bib}

\end{document}